\documentclass{article}
\usepackage{ijcai19}

\usepackage{times}
\usepackage{soul}
\usepackage{color}
\usepackage{caption}
\usepackage{url}
\usepackage[hidelinks]{hyperref}
\usepackage[utf8]{inputenc}
\usepackage[font=small]{caption}
\usepackage{graphicx}
\usepackage{amsmath}
\usepackage{booktabs}
\usepackage{float}
\usepackage{algorithm}
\usepackage{algorithmic}

\newtheorem{theorem}{Theorem}
\title{Boosting Generative Models by Leveraging Cascaded Meta-Models}

\author{Fan Bao
\and Hang Su
\and Jun Zhu\\
\vspace{0.1cm}
\textnormal{\small{Institute for Artificial Intelligence\\
State Key Laboratory of Intelligent Technology and Systems\\
Beijing National Research Center for Information Science and Technology\\
Department of Computer Science and Technology, Tsinghua University, Beijing 100084, China}}
}

\begin{document}

\maketitle

\begin{abstract}
  Deep generative models are effective methods of modeling data. However, it is not easy for a single generative model to faithfully capture the distributions of complex data such as images. In this paper, we propose an approach for boosting generative models, which cascades meta-models together to produce a stronger model. Any hidden variable meta-model (e.g., RBM and VAE) which supports likelihood evaluation can be leveraged. We derive a decomposable variational lower bound of the boosted model, which allows each meta-model to be trained separately and greedily. Besides, our framework can be extended to semi-supervised boosting, where the boosted model learns a joint distribution of data and labels. Finally, we combine our boosting framework with the multiplicative boosting framework, which further improves the learning power of generative models.
\end{abstract}

\section{Introduction}


The past decade has witnessed tremendous success in the field of deep generative models (DGMs) 
in both unsupervised learning \cite{goodfellow2014generative,kingma2013auto,radford2015unsupervised} and semi-supervised learning \cite{abbasnejad2017infinite,kingma2014semi,li2018max} paradigms. DGMs learn the data distribution by combining the scalability of deep learning with the generality of probabilistic reasoning. However, it is not easy for a single parametric model to learn a complex distribution, since the upper limit of a model ability is determined by its fixed structure. If a model with low capacity was adopted, the model would be likely to have a poor performance. Straightforwardly increasing the model capacity (e.g., including more layers or more neurons) is likely to cause serious challenges, such as vanishing gradient problem \cite{hochreiter2001gradient} and exploding gradient problem \cite{grosse2017lecture}. 

An alternative approach is to integrate multiple weak models to achieve a strong one. The early success was made on mixture models~\cite{dempster1977maximum,figueiredo2002unsupervised,xu1996convergence} and product-of-experts~\cite{hinton1999products,hinton2002training}. However, the weak models in such work are typically shallow models with very limited capacity. Recent success has been made on   
boosting generative models, where a set of meta-models (i.e., weak learners) are combined to construct a stronger model. In particular, \cite{grover2018boosted} propose a method of multiplicative boosting, which takes the geometric average of the meta-model distributions, with each assigned an exponentiated weight.
This boosting method improves performance on density estimation and sample generation, compared to a single meta-model. However, the boosted model has an explicit partition function, which requires importance sampling~\cite{rubinstein2016simulation}. In general, sampling from the boosted model is conducted based on Markov chain Monte Carlo (MCMC) method~\cite{hastings1970monte}. As a result, it requires a high time complexity of likelihood evaluation and sample generation. Rosset \textit{et al.}~\shortcite{rosset2003boosting} propose another method of additive boosting, which takes the weighted arithmetic mean of meta-models' distributions. This method can sample fast, but the improvement of performance on density estimation is not comparable to the multiplicative boosting, since the additive boosting requires that the expected log-likelihood and likelihood of the current meta-model are better-or-equal than those of the previous boosted model, which is difficult to satisfy~\cite{grover2018boosted}.
In summary, it is nontrivial for both of the previous boosting methods to balance well between improving the learning power and keeping the efficiency of sampling and density estimation. 


To address the aforementioned issues, we propose a novel boosting framework, where meta-models are connected in cascade. Our meta-algorithmic framework is inspired by the greedy layer-wise training algorithm of DBNs (Deep Belief Networks)~\cite{bengio2007greedy,hinton2006fast}, where an ensemble of RBMs (Restricted Boltzmann Machines)~\cite{smolensky1986information} are converted to a stronger model. We propose a decomposable variational lower bound, which reveals the principle behind the greedy layer-wise training algorithm. The decomposable lower bound allows us to incorporate any hidden variable meta-model  (e.g. RBM and VAE (Variational Autoencoder)~\cite{kingma2013auto}), as long as it supports likelihood evaluation, and to train these meta-models separately and greedily, yielding a deep boosted model.
Besides, our boosting framework can be extended to semi-supervised boosting, where the boosted model learns a joint distribution of data and labels. Finally, We demonstrate that our boosting framework can be integrated with the multiplicative boosting framework~\cite{grover2018boosted}, yielding a hybrid boosting with an improved learning power of generative models.
To summary, we make the following contributions:
\begin{itemize}
    \item We propose a meta-algorithmic framework to boost generative models by cascading the hidden variable meta-models, which can be also extended to semi-supervised boosting.
    \item We give a decomposable variational lower bound of the boosted model, which reveals the principle behind the greedy layer-wise training algorithm.
    \item We finally demonstrate that our boosting framework can be extended to a hybrid model by integrating it with the multiplicative boosting models, which further improves the learning power of generative models.
\end{itemize}

\section{Approach}
\label{sec:boosting}
In \autoref{sec:mul_boosting}, we review the current multiplicative boosting \cite{grover2018boosted}. 
Then, we present our boosting framework. We first figure out how to connect meta-models, and then propose our meta-algorithmic framework with its theoretical analysis. Afterwards, we discuss the convergence and extension to semi-supervised boosting.

\subsection{Boosting Generative Models}
\label{sec:mul_boosting}
Grover and Ermon \shortcite{grover2018boosted} introduced multiplicative boosting, which takes the geometric average of meta-models' distributions, with each assigned an exponentiated weight $\alpha_i$ as 
\begin{equation}
    P_n = \frac{\prod_{i=0}^n M_i^{\alpha_i}}{Z_n},
\end{equation}
where $M_i$ ($0\leq i \leq n$) are meta-models, which are required to support likelihood evaluation, $P_n$ is the boosted model and $Z_n$ is the partition function. The first meta-model $M_0$ is trained on the empirical data distribution $p_D$ which is defined to be uniform over the dataset $D$. The other meta-models $M_i$ ($1\leq i \leq n$) are trained on a reweighted data distribution $p_{D_i}$ as 
\begin{equation}
\label{eq:mul_boosting}
    \max_{M_i}\mathbf{E}_{p_{D_i}}\left[\mathrm{log}M_i\right],
\end{equation}
where $\ p_{D_i} \propto \left(\frac{p_D}{P_{i-1}}\right)^{\beta_i}$ with 
$\beta_i \in \left[0, 1\right]$ being the hypermeter. These meta-models are therefore connected in parallel, as shown in \autoref{fig:connection_comp}, since their distributions are combined by multiplication.

Grover and Ermon \shortcite{grover2018boosted} show that the expected log-likelihood of the boosted model $P_i$ over the dataset $D$ will not decrease (i.e., $\mathbf{E}_{p_D}\left[\mathrm{log}P_i\right]\geq \mathbf{E}_{p_D}\left[\mathrm{log}P_{i-1}\right]$) if \autoref{eq:mul_boosting} is maximized. The multiplicative boosting makes success in improving the learning power power of generative models. Compared to an individual meta-model, it has better performance on density estimation and generates samples of higher quality. However, importance sampling and MCMC are required to evaluate the partition function $Z_n$ and generate samples respectively, which limits its application in occasions requiring fast density estimation and sampling. To overcome these shortcomings, we propose our boosting framework, where meta-models are cascaded.

\subsection{How to Connect Meta-Models}
\label{sec:connection}
\begin{figure}[ht]\vspace{-.4cm}
\begin{center}
\centerline{\includegraphics[width=\columnwidth]{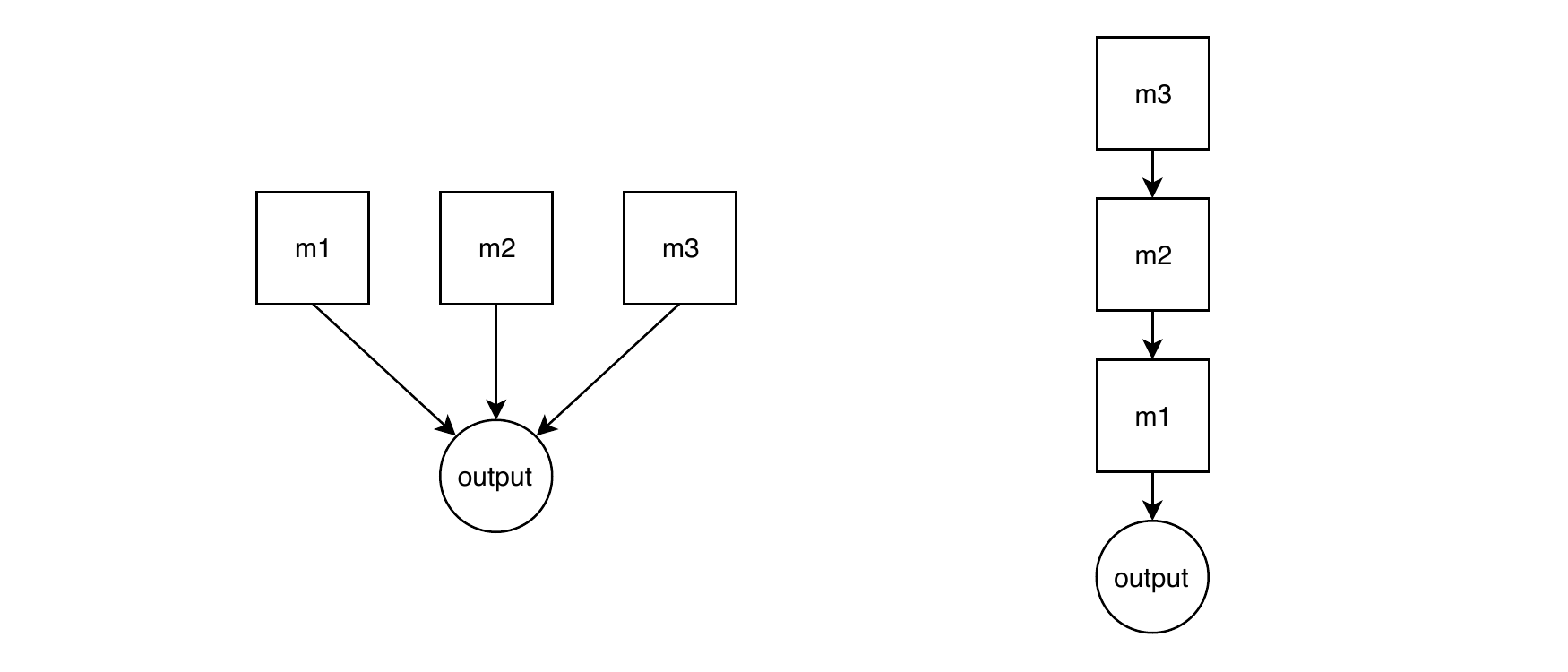}}\vspace{-.4cm}
\caption{\textbf{Left:} the parallel connection of meta-models in the multiplicative boosting framework. \textbf{Right:} the cascade connection of meta-models in our boosting framework.}
\label{fig:connection_comp}
\end{center}\vspace{-.4cm}
\end{figure}

In multiplicative boosting, meta-models are connected in parallel, leading to the troublesome partition function. To overcome this problem, we connect meta-models in cascade, where the output of a previous model is passed to the input of the next model, as shown in \autoref{fig:connection_comp}.

Now let's formulate our cascade connection for generative models. Suppose we have a group of meta-models $m_i(x_i, h_i)$, $i=1, \cdots, n$, where $x_i$ is the visible variable and $h_i$ is the hidden variable. These models can belong to different families (e.g. RBMs and VAEs), as long as they have hidden variables and support likelihood evaluation. 
To ensure the output of the previous model is passed to the input of the next model, we replace $x_i$ with $h_{i-1}$ and connect meta-models in a directed chain:
\begin{equation}
    p_k(x, h_1, \cdots, h_k) = m_k(h_{k-1}, h_k)\prod_{i=1}^{k-1}m_i(h_{i-1}|h_i),
\end{equation}
where $h_0=x$ and $k\leq n$. The visible variable of a previous model severs as the hidden variable of the next model, so that we can sample the boosted model fast in a top-down style. It is worth note that, when all meta-models are RBMs, the boosted model is a DBN and when all meta-models are VAEs, the boosted model is a DLGM (Deep Latent Gaussian Model)~\cite{burda2015importance,rezende2014stochastic}.

The boosted model allows us to implement generating hereby. Then, we build the approximation of the posterior distribution, which allows to do inference. We also connect these meta-models in a directed chain \cite{burda2015importance},
so that we can do inference in a down-top style as
\begin{equation}
\label{eq:post}
    q(h_1, \cdots, h_k|x)=\prod_{i=1}^k m_i(h_i|h_{i-1}),
\end{equation}
where $h_0 = x$ and $k\leq n$. The approximation of the posterior distribution has the advantage that we don't need to re-infer the whole boosted model after incorporating a new meta-model $m_{k}$: we only need to infer $h_{k}$ from $m_{k}(h_{k}|h_{k-1})$ conditioned on $h_{k-1}$ inferred previously.


\subsection{Decomposable Variational Lower Bound}
\label{sec:elbo}
Suppose $D$ is our training dataset, we give a lower bound to the marginal likelihood $\mathbf{E}_D \left[\mathrm{log}p_k(x)\right]$ of the boosted model $p_k$. The lower bound is decomposed to $k$ terms, which reveals the principle behind the greedy layer-wise training algorithm, as given in \autoref{thm:ELBO}.
\begin{theorem}
\label{thm:ELBO}
Let $m_i(h_{i-1}, h_i)$($1 \leq i \leq n$) be meta-models, $p_k(x, h_1, \cdots, h_k) = m_k(h_{k-1}, h_k)\prod_{i=1}^{k-1}m_i(h_{i-1}|h_i)$ be the model boosted from meta-models $m_i$($1\leq i\leq k$), $q(h_1, \cdots, h_{k-1}|x)=\prod_{i=1}^{k-1} m_i(h_i|h_{i-1})$ be the approximate posterior, then we have:
\begin{equation}
\label{eq:elbo}
\begin{aligned}
    &\mathbf{E}_D\left[\mathrm{log}p_{k}(x)\right] \geq\\  
    &\mathbf{E}_D\mathbf{E}_{q(h_1, \cdots, h_{k-1}|x)} \left[ \mathrm{log} \frac{p_{k}(x, h_1, \cdots, h_{k-1})}{q(h_1, \cdots, h_{k-1}|x)} \right]= \sum_{i=1}^{k} \mathcal{L}_k,
\end{aligned}
\end{equation}
where $\mathcal{L}_1 = \mathbf{E}_D\left[\mathrm{log}m_1(x)\right]$ and
\begin{equation}
\begin{aligned}
        \mathcal{L}_i=&\mathbf{E}_D\mathbf{E}_{q(h_{i-1}|x)} \left[\mathrm{log}m_{i}(h_{i-1})\right]\\
    &-\mathbf{E}_D\mathbf{E}_{q(h_{i-1}|x)}\left[\mathrm{log}m_{i-1}(h_{i-1})\right] (2\leq i \leq k).
\end{aligned}
\end{equation}
Proof: see \rm{Appendix A}\footnote{The appendix is in supplemental material at \url{https://github.com/nspggg/BMG_Cascade/blob/master/SupplementalMaterial.pdf}.\label{ft:appendix}}
\end{theorem}

The lower bound is decomposed to $k$ terms $\mathcal{L}_i$ ($1\leq i \leq k$). Specifically, $\mathcal{L}_1$ is the marginal likelihood of the first meta-model $m_1$. $\mathcal{L}_i$ ($2\leq i \leq k$) is the difference between the marginal likelihood of the observable variable for $m_i$ and the hidden variable for $m_{i-1}$.

When $k$ = 1, there is only one meta-model and the lower bound is exactly equal to the marginal likelihood of the boosted model. So the lower bound is tight when $k=1$. Based on the initially tight lower bound, we can further promote it by optimizing these decomposed terms sequentially, yielding the greedy layer-wise training algorithm, as discussed in \autoref{sec:train}.

\subsection{The Greedy Layer-Wise Training Algorithm}
\label{sec:train}
The difference between the lower bound of $\mathbf{E}_D \left[\mathrm{log}p_{k}(x)\right]$ and $\mathbf{E}_D\left[ \mathrm{log}p_{k-1}(x)\right]$ is:
\begin{equation}
\label{eq:diff}
\begin{aligned}
    \mathcal{L}_{k}=&\mathbf{E}_D\mathbf{E}_{q(h_{k-1}|x)}\left[\mathrm{log}m_{k}(h_{k-1})\right]\\
    &-\mathbf{E}_D\mathbf{E}_{q(h_{k-1}|x)}\left[\mathrm{log}m_{k-1}(h_{k-1})\right].
\end{aligned}
\end{equation}
To ensure the lower bound grows with $m_{k}$ incorporated, we only need to ensure $\mathcal{L}_k$ is positive. When we train the meta-model $m_{k}$, we fix rest meta-models $m_i$ ($1\leq i \leq k-1$) and optimize $\mathbf{E}_D\mathbf{E}_{q(h_{k-1}|x)}\left[\mathrm{log}m_{k}(h_{k-1})\right]$ by only tuning the parameters of $m_{k}$ until it is greater than $\mathbf{E}_D\mathbf{E}_{q(h_{k-1}|x)}\left[\mathrm{log}m_{k-1}(h_{k-1})\right]$. As a result, we can train each meta-model separately and greedily, as outlined in Alg.~\ref{alg:train}. 
\begin{algorithm}[t]
    \caption{Boosting Generative Models}
    \label{alg:train}
\begin{algorithmic}[1]
    \STATE {\bfseries Input:} dataset $D$;\ \ number of meta-models $n$
    \STATE Let $p_D=\sum_{x^{(i)}\in D}\delta(x-x^{(i)})/|D|$ be the empirical distribution of $D$
    \STATE Sample from $p_D(x)$ and train $m_1(x, h_1)$
    \STATE $k=2$
    \WHILE{$k \leq n$}
        \STATE $q(h_1, \cdots, h_{k-1}|x)=\prod_{i=1}^{k-1} m_i(h_i|h_{i-1})$
        \STATE Sample $h_{k-1}$ from $p_D(x)q(h_{k-1}|x)$
        \STATE Use these samples to train $m_{k}(h_{k-1}, h_{k})$ 
        \STATE $k= k+1$
    \ENDWHILE
    \STATE $p_n(x, h_1, \cdots, h_n)= m_n(h_{n-1},h_n)\mathop{\prod}\limits_{i=1}^{n-1}m_i(h_{i-1}|h_i)$
    \STATE \textbf{return} $p_n(x, h_1, \cdots, h_n)$
\end{algorithmic}
\end{algorithm}

Generally, after training $m_{k}$, \autoref{eq:diff} will not be negative and the lower bound in \autoref{eq:elbo} will not decrease, since if $m_{k}$ is an arbitrarily powerful learner, we can make sure~\autoref{eq:diff} is greater-or-equal than zero:
\begin{equation}
\begin{aligned}
    \mathbf{E}_D\mathbf{E}_{q(h_{k-1}|x)}&\left[\mathrm{log}m_{k}(h_{k-1})\right]\\
    &=\mathop{\mathrm{max}}\limits_p \mathbf{E}_D\mathbf{E}_{q(h_{k-1}|x)}\left[\mathrm{log}p(h_{k-1})\right]\\
    &\geq \mathbf{E}_D\mathbf{E}_{q(h_{k-1}|x)}\left[\mathrm{log}m_{k-1}(h_{k-1})\right].
\end{aligned}
\end{equation}
In practice, \autoref{eq:diff} is likely to be negative under the following three cases:
\begin{itemize}
    \item $m_{k}$ is not well trained. In this case, \autoref{eq:diff} is very negative, which indicates us to tune hyperparameters of $m_{k}$ and retrain this meta-model.
    \item $m_{k-1}(h_{k-1})$ is close to the marginal distribution of $p_D(x) q(h_{k-1}|x)$. In this case, \autoref{eq:diff} will be close to zero, and we can either keep training by incorporating more powerful meta-models or just stop training.
    \item The lower bound converges. In this case, the lower bound will stop growing even if you keep incorporating more meta-models. The convergence will be further discussed in \autoref{sec:convergence}
\end{itemize}
For models with $m_{k}(h_{k-1})$ initialized from $m_{k-1}(h_{k-1})$, such as DBNs~\cite{hinton2006fast}, we can make sure that ~\autoref{eq:diff} will never be negative. 


\subsection{Convergence}
\label{sec:convergence}
It's impossible for the decomposable lower bound to grow infinitely. After training $m_k$, if it reaches the best global optimum (i.e., $m_k(h_{k-1})$ is exactly the marginal distribution of $p_D(x)q(h_{k-1}|x)$), then the lower bound will stop growing even if we keep incorporating more meta-models. We formally describe the convergence in \autoref{thm:Convergence}.

\begin{theorem}
\label{thm:Convergence}
If $m_k$ reaches the best global optimum, then $\sum_{i=k+1}^j \mathcal{L}_i \leq 0$, for any $j \geq k + 1$.

Proof: see \rm{Appendix B\textsuperscript{\ref{ft:appendix}}}.
\end{theorem}

It indicates us that it is unnecessary to incorporate meta-models as much as possible, since the boosted model will converge if $m_k$ reaches the best global optimum. We give another necessary condition of the convergence in \autoref{thm:Convergence_necc}.

\begin{theorem}
\label{thm:Convergence_necc}
If $m_k$ reaches the best global optimum, then $\mathbf{E}_D\mathbf{E}_{q(h_{k}|x)}\left[\mathrm{log}m_{k}(h_k)\right]=\mathbf{E}_{m_k(h_{k})}\left[\mathrm{log}m_{k}(h_k)\right]$.

Proof: see \rm{Appendix B\textsuperscript{\ref{ft:appendix}}}.
\end{theorem}

For meta-models such as VAEs, $m_k(h_k)$ is the standard normal distribution and $\mathbf{E}_{m_k(h_{k})}\left[\mathrm{log}m_{k}(h_k)\right]$ is analytically solvable, which helps us judge whether the boosted model has converged in practice.

\subsection{Semi-Supervised Extension}
\label{sec:semi}
Our approach can be extended to semi-supervised learning, where the joint distribution is
\begin{equation}
    p_n(x, y, h_1, \cdots, h_n) = m_n(h_{n-1},y, h_n)\prod_{i=1}^{n-1}m_i(h_{i-1}|h_i),\nonumber
\end{equation}
and the approximation of posterior is chosen as
\begin{equation}
    q(y, h_1, \cdots, h_n|x)=m_n(y, h_n|h_{n-1})\prod_{i=1}^{n-1} m_i(h_i|h_{i-1}),\nonumber
\end{equation}
where the latent class variable $y$ only appears in the last meta-model $m_n$. The model is also trained greedily and layer-wise, with the first $n-1$ meta-models trained on unlabelled data using Alg.~\ref{alg:train}, yielding the latent representations $h_{n-1}$ of data. The top-most meta-model $m_n$ is trained on both unlabelled and labelled data, where non-labelled parts are first converted to their latent representations. 
The formal training algorithm with its detailed derivation and experiments are given in Appendix C\textsuperscript{\ref{ft:appendix}}. The experiments show that our semi-supervised extension can achieve good performance on the classification task without the help of any classification loss.


\section{Hybrid Boosting}
\label{sec:hybrid}


Now we know that meta-models can be  connected either in cascade, as given by our boosting method, or in parallel, as given by the multiplicative boosting; we can further consider the hybrid boosting, where a boosted model contains both cascade and parallel connections. In fact, it is not difficult to implement: we can think of the boosted model produced by our method as the meta-model for multiplicative boosting.

An open problem for hybrid boosting is to determine what kind of meta-models to use and how meta-models are connected, which is closely related to the specific dataset and task. Here we introduce some strategies for this problem. 

For cascade connection, if the dataset can be divided to several categories, it is appropriate to use a GMM (Gaussian Mixture Model) \cite{smolensky1986information} as the top-most meta-model. Other meta-models can be selected as VAEs \cite{kingma2013auto} or their variants \cite{burda2015importance,sonderby2016ladder}. There are three reasons for this strategy: (1) the posterior of VAE is much simpler than the dataset distribution, making a GMM enough to learn the posterior; (2) the posterior of a VAE is likely to consist of several components, with each corresponding to one category, making a GMM which also consists of several components suitable; (3) Since $m_{k-1}(h_{k-1})$ is a standard Gaussian distribution when $m_{k-1}$ is a VAE, according to \autoref{eq:diff}, if $m_{k}(h_{k-1})$ is a GMM, which covers the standard Gaussian distribution as a special case, we can make sure that \autoref{eq:diff} will not be negative after training $m_{k}$. 

For parallel connection, each meta-model should have enough learning power for the dataset, since each meta-model is required to learn the distribution of the dataset or the reweighted dataset. If any meta-model fails to learn the distribution, the performance of the boosted model will be harmed. In \autoref{sec:comp_boosting}, we give a negative example, where a VAE and a GMM are connected in parallel and the overall performance is extremely bad.

\section{Experiments}
We now present experiments to verify the effectiveness of our method. We first give results of boosting a set of RBMs and VAEs to validate that the performance of the boosted model is really promoted with more meta-models incorporated. Then, we give results of boosting some advanced models to show that our method can be used as a technique to further promote the performance of state-of-the-art models. Finally, we make comparison between different generative boosting methods.

\subsection{Setup}
We do experiments on binarized mnist~\cite{lecun-mnisthandwrittendigit-2010}, which contains 60000 training data and 10000 testing data, as well as the more complex celebA dataset~\cite{liu2015faceattributes}, which contains 202599 face images, with each first  resized to $64\times 64$. The meta-models we use are RBMs~\cite{smolensky1986information}, GMMs~\cite{reynolds2015gaussian}, VAEs~\cite{kingma2013auto}, ConvVAEs (i.e., VAEs with convolutional layers), IWAEs~\cite{burda2015importance}, and  LVAEs~\cite{sonderby2016ladder}, with their architectures given in Appendix D\textsuperscript{\ref{ft:appendix}}. All experiments are conducted on one 2.60GHz CPU and one GeForce GTX.

\subsection{Boosting RBMs and VAEs}
\label{sec:boosting_vaes_rbms}
Using RBMs and VAEs as meta-models, we first evaluate the lower bound of boosted models, and then generate samples from them. Finally, we compare our boosting method with the method of naively increasing model capacity. 

\begin{figure}[htb]\vspace{-.2cm}
  \centering
    \includegraphics[width=1.0\linewidth]{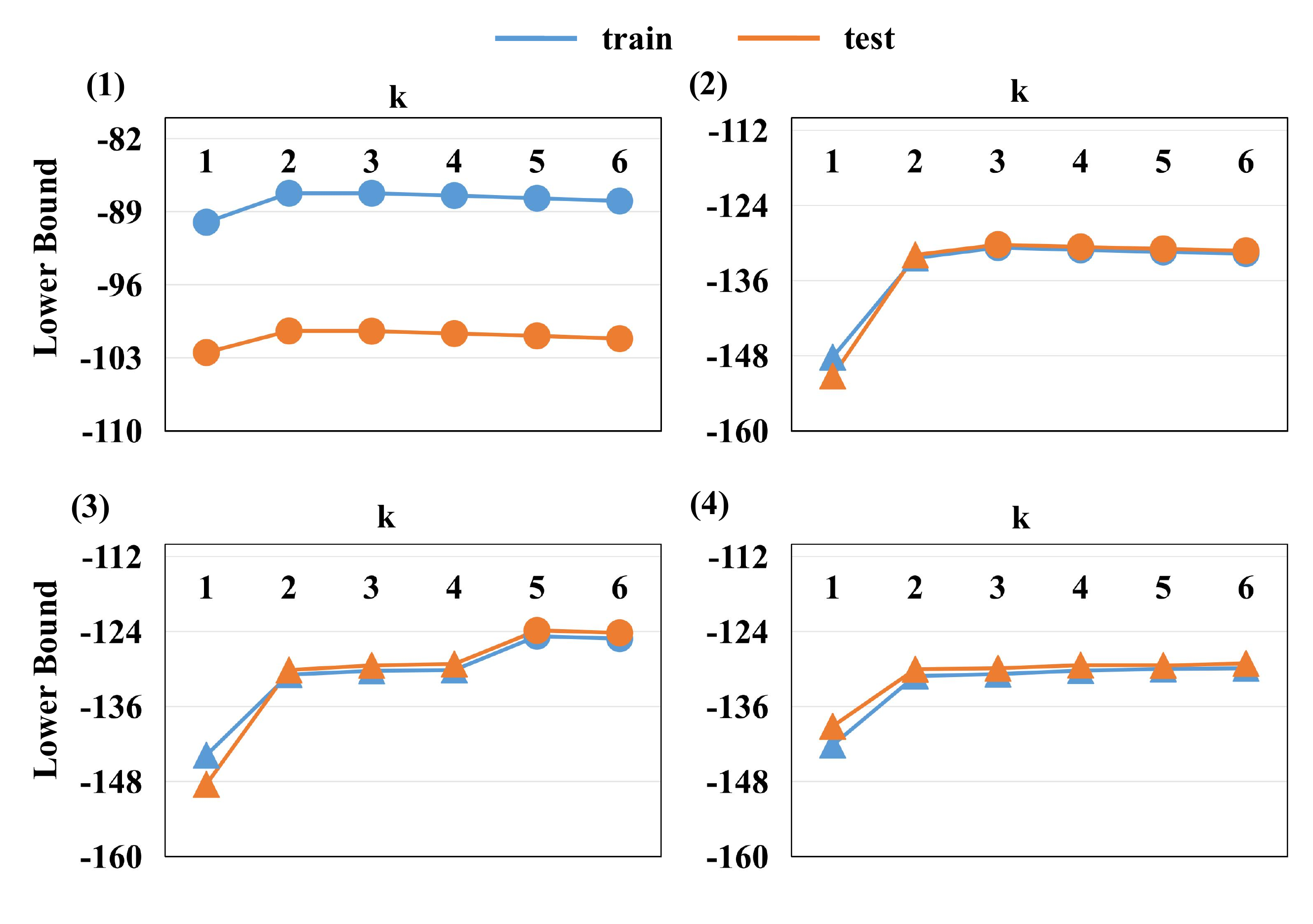}\vspace{-.3cm}
     \captionsetup{font={footnotesize}}
    \caption{The lower bound (\autoref{eq:elbo}) on different combinations of meta-models. The triangular and circular markers correspond to RBMs and VAEs respectively. \textbf{(1):} All meta-models are VAEs. After incorporating two VAEs, the lower bound enters a stable state. 
    \textbf{(2):} The first two meta-models are RBMs and the rest are VAEs. The second RBM greatly promotes the lower bound and the first VAE helps slightly.
    \textbf{(3):} The first four meta-models are RBMs and the rest are VAEs. The lower bound grows as the first two RBMs are incorporated, while the incorporation of next two RBMs doesn't help promote the lower bound. We further improve the lower bound by adding two VAEs. 
    \textbf{(4):} All meta-models are RBMs. After incorporating two RBMs, the lower bound enters a stable state.
    }
    \label{fig:elbo}\vspace{-.3cm}
\end{figure}

\paragraph{Evaluation of lower bound.} Firstly, we evaluate the lower bound on 4 combinations of RBMS and VAEs on mnist, where a higher lower bound corresponds to a better performance. Since the stochastic variables in RBMs are discrete and the stochastic variables in VAEs are continuous, we put RBMs at bottom and put VAEs at top. For each combination, we evaluate the lower bound in \autoref{eq:elbo} at different $k$ $(1\leq k \leq 6)$ on both mnist training and testing dataset. The result is shown in \autoref{fig:elbo}, where the triangular and circular markers correspond to RBMs and VAEs respectively. If we keep adding models of the same kind, the lower bound will first increases and then reachs a stable state, as shown in Chart (1) and Chart (4). But it doesn't mean that the lower bound has converged to the best state and will not increase. As shown in Chart (3), the boosted model reaches a stable state after incorporating two RBMs, but we can further improve the lower bound by incorporating VAEs.


\begin{figure}[htb]
  \centering
    \includegraphics[width=1.0\linewidth]{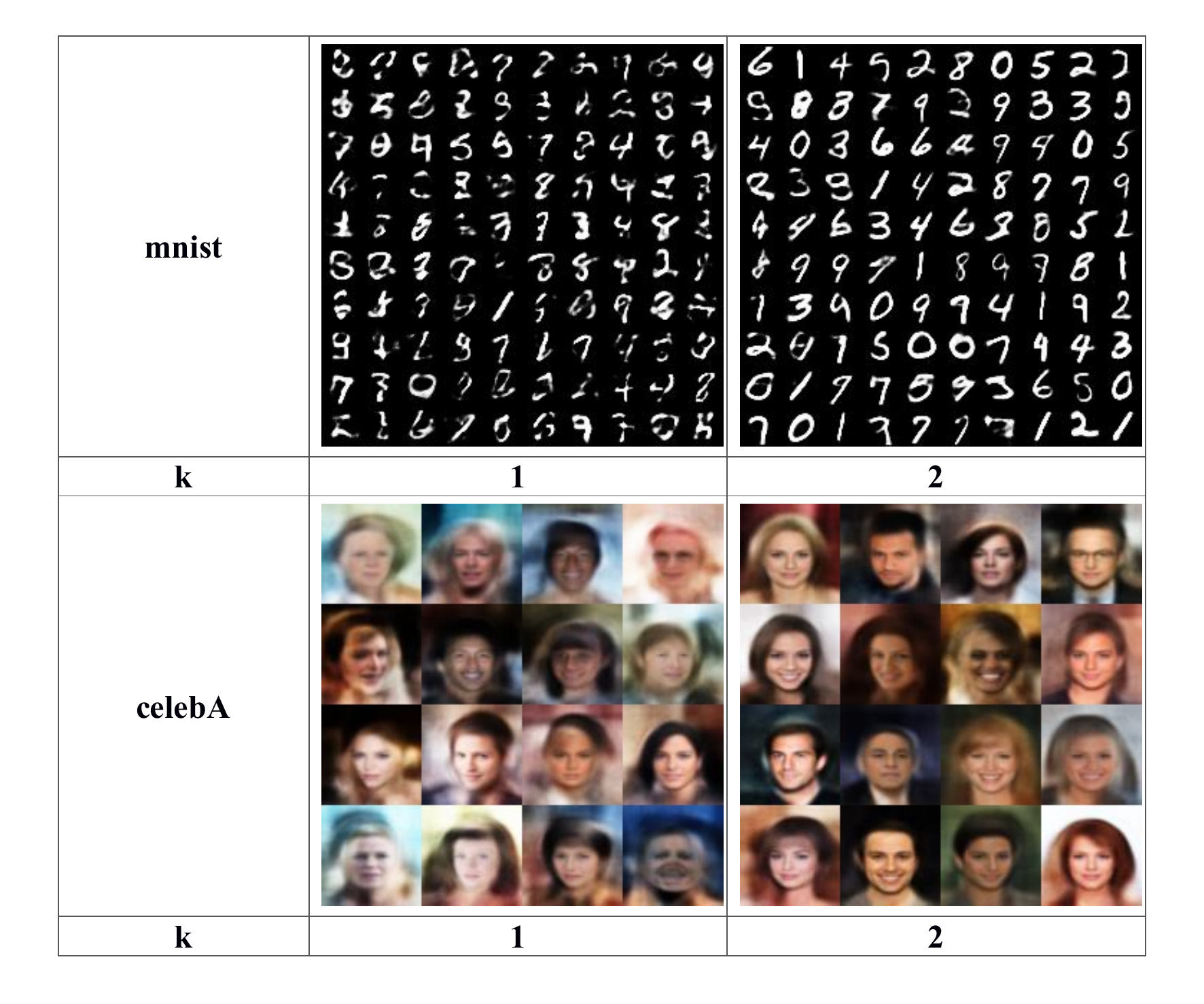}\vspace{-.3cm}
     \captionsetup{font={footnotesize}}
    \caption{Samples generated from boosted models consisting of different number of VAEs. k is the number of VAEs.} 
    \label{fig:VAE_mnist}\vspace{-.3cm}
\end{figure}

\paragraph{Sample generation.} We sample randomly from boosted models consisting of one and two VAEs respectively. As shown in \autoref{fig:VAE_mnist}, the incorporation of an extra VAE increases the quality of generated images for both mnist and celebA, which is consistent with the performance on density estimation over the two datasets, as shown in \autoref{tab:density_est_dataset}.
\begin{table}[H]
\centering
 \captionsetup{font={footnotesize}}

\begin{tabular}{|l|r|r|}  
	\hline
	$k$ (Number of VAEs) & mnist & celebA\\
	\hline
    $k=1$ & -102.51 & -6274.86\\
    \hline
    $k=2$ & -100.44 & -6268.43\\
    \hline
\end{tabular}
\caption{Density estimation over mnist test set and celebA.}
\label{tab:density_est_dataset}
\end{table}

\paragraph{Comparison with naively increasing model capacity.} We compare our methods with the method of naively increasing model capacity. The conventional method of increasing model capacity is either to add more deterministic hidden layers or to increase the dimension of deterministic hidden layers, so we compare our boosted model (Boosted VAEs) with a deeper model (Deeper VAE) and a wider model (Wider VAE). The Deeper VAE has ten 500-dimensional deterministic hidden layers; the Wider VAE has two 2500-dimensional deterministic hidden layers; the Boosted VAEs is composed of 5 base VAEs, each of them has two 500-dimensional deterministic hidden layers. As a result, all the three models above have 5000 deterministic hidden units. 

\begin{figure*}[htb]
  \centering
    \includegraphics[width=1.0\linewidth]{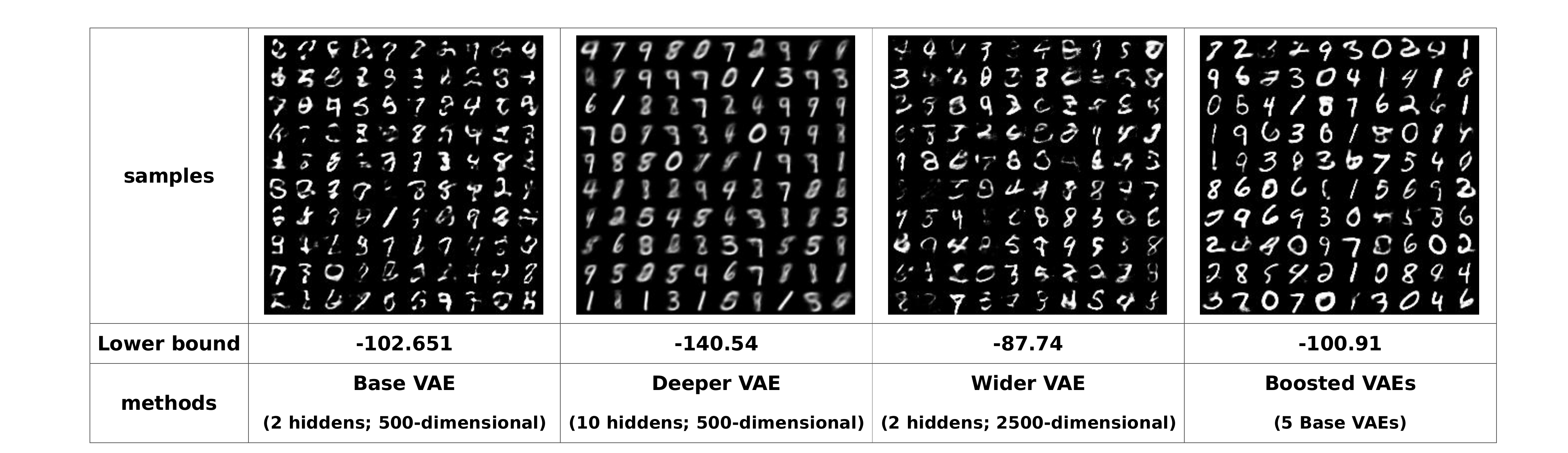}\vspace{-.3cm}
     \captionsetup{font={footnotesize}}
    \caption{Comparison between our boosting method and the method of naively increasing model capacity. Deeper VAE, Wider VAE and Boosted VAEs have the same number of deterministic hidden units. Wider VAE has the highest lower bound, but most digits generated by Wider VAE are undistinguishable. Meanwhile, the Deeper VAE is able to generate distinguishable digits, but some digits are rather blurred and its lower bound is the lowest one. Only the digits generated by Boosted VAEs are both distinguishable and sharp.}
    \label{fig:longVAE}
\end{figure*}

\autoref{fig:longVAE} shows the results. Wider VAE has the highest lower bound, but its generated digits are usually undistinguishable. Meanwhile, the Deeper VAE is able to generate distinguishable digits, but some digits are rather blurred and its lower bound is the lowest one. Only the digits generated by Boosted VAEs are both distinguishable and sharp. 

Since straightforwardly increasing the model capacity is likely to cause serious challenges, such as vanishing gradient problem \cite{hochreiter2001gradient} and exploding gradient problem \cite{grosse2017lecture}, it often fails to achieve the desired results on improving models' learning power. Our boosting method avoids these challenges and achieves better result than the method of naively increasing the model capacity.

\subsection{Boosting Advanced Models}
\label{sec:boosting_adv_models}
We choose ConvVAE (i.e., VAE with convolutional layers), LVAE~\cite{sonderby2016ladder}, IWAE~\cite{burda2015importance} as advanced models, which represent current state-of-art methods. We use one advanced model and one GMM~\cite{reynolds2015gaussian} to construct a boosted model, with the advanced model at the bottom and the GMM at the top. The result is given in \autoref{tab:boosting_adv_models}. We see that the performance of each advanced model is further increased by incorporating a GMM, at the cost of a few seconds. 

The performance improvement by incorporating a GMM is theoretically guaranteed: since $m_{1}(h_{1})$ is a standard Gaussian distribution when $m_1$ is a VAE or one of the above three advanced variants, according to \autoref{eq:diff}, if $m_{2}(h_{1})$ is a GMM, which covers the standard Gaussian distribution as a special case, we can make sure that $\mathcal{L}_2$ will not be negative after training $m_{2}$. Besides, the dimension of hidden variable $h_1$ is much smaller than the dimension of observable variable $h_0$ for VAEs and their variants, and thus the training of $m_2$ only requires very little time. As a result, our boosting framework can be used as a technique to further promote the performance of state-of-the-art models, at the cost of very little time.

\begin{table}[H]\vspace{-.2cm}
\centering 
 \captionsetup{font={footnotesize}}

\begin{tabular}{|l|r|r|}  
	\hline
	   & $\leq \mathrm{log}p(x)$ & extra time (s)\\
	\hline
    ConvVAE & -88.41 & \\
    ConvVAE + GMM & \textbf{-87.42} & +7.85\\
    \hline
    LVAE, 2-layer & -95.73 & \\
    LVAE, 2-layer + GMM & \textbf{-95.50} &+11.76 \\
    \hline
    IWAE, k=5 & -81.58 & \\
    IWAE, k=5 + GMM & \textbf{-80.38} & +9.41\\
    \hline
    IWAE, k=10 & -80.56 & \\
    IWAE, k=10 + GMM & \textbf{-79.20} & +8.39 \\
    \hline
\end{tabular}\vspace{-.2cm}
\caption{Test set performance on mnist. LVAE has 2 stochastic hidden layers. The number of importance weighted samples (k) for IWAE is 5 and 10. The number of components in GMM is set to 10. The extra time is the time cost for incorporating an extra GMM.}
\label{tab:boosting_adv_models}\vspace{-.4cm}
\end{table}

\begin{table*}[htbp]
\centering 
 \captionsetup{font={footnotesize}}
\begin{tabular}{|l|l|r|r|r|r|}  
	\hline
	 & & $\leq \mathrm{log}p(x)$ & training time (s) & density estimation time (s) & sampling time (s)\\
	\hline
    cascade & VAE+VAE & -99.53 & 223.85 & 0.42 & 0.13\\
    & VAE+GMM & -98.13 & 116.33 & 0.14 & 0.12\\
    \hline
    parallel & VAE$\parallel$VAE & -95.72 & 225.21 & 50.91 & 543.82\\
    & VAE$\parallel$GMM & -506.76 & 2471.60 & 130.65 & 480.95\\
    \hline
    hybrid & (VAE+GMM)$\parallel$VAE & -94.28 & 225.23 & 125.20 & 1681.77\\
    & (VAE+GMM)$\parallel$(VAE+GMM) & -93.94 & 226.86 & 147.82 & 2612.59\\
    \hline
\end{tabular}\vspace{-.2cm}
\caption{Comparison between different boosting methods on mnist. The `+' represents the cascade connection and the `$\parallel$' represents the parallel connection. The density is estimated on the test set and the sampling time is the time cost for sampling 10000 samples.}
\label{tab:bybrid_boosting}\vspace{-.3cm}
\end{table*}

\subsection{Comparison between Different Generative Boosting Methods}
\label{sec:comp_boosting}
We compare our boosting framework where meta-models are connected in cascade, multiplicative boosting where meta-models are connected in parallel, and hybrid boosting where meta-models are connected in both cascade and parallel. The result is given in \autoref{tab:bybrid_boosting}. The hybrid boosting produces the strongest model, but the time cost of density estimation and sampling is high, since the structure includes parallel connections. The cascade connection allows quick density estimation and sampling, but the boosted model is not as strong as the hybrid boosting. It is also worth note that the parallel connection of one VAE and one GMM produces a bad model, since the learning power of a GMM is too weak for mnist dataset and the training time of a GMM is long for high dimensional data.

\section{Related Work}
\paragraph{Deep Belef Networks.} Our work is inspired by DBNs~\cite{hinton2006fast}. A DBN has a multi-layer structure, whose basic components are RBMs~\cite{smolensky1986information}.  During training, each RBM is learned separately, and stacked to the top of current structure. It is a classical example of boosting generative model, since a group of RBMs are connected to produce a stronger model. 
Our decomposable variational lower bound reveals the principle behind the training algorithm of DBNs: since for DBNs, $m_{k}(h_{k-1})$ is initialized from $m_{k-1}(h_{k-1})$, we can make sure that \autoref{eq:diff} will never be negative and the lower bound will never decrease.

\paragraph{Deep Latent Gaussian Models.} DLGMs are deep directed graphical models with multiple layers of hidden variables~\cite{burda2015importance,rezende2014stochastic}. The distribution of hidden variables in layer $k$ conditioned on hidden variables in layer $k+1$ is a Gaussian distribution. Rezende \textit{et al.}~\shortcite{rezende2014stochastic} introduce an approximate posterior distribution which factorises across layers. Burda \textit{et al.}~\shortcite{burda2015importance} introduce an approximate posterior distribution which is a directed chain. When we restrict our meta-models to VAEs, we derive the same variational lower bound to Burda \textit{et al.}~\shortcite{burda2015importance}. The difference is that Burda \textit{et al.}~\shortcite{burda2015importance} optimize the lower bound as a whole, but our work optimizes the lower bound greedily and layer-wise.


\paragraph{Other methods of boosting generative models.} Methods of boosting generative models have been explored. Previous work can be divided into two categories: \textit{sum-of-experts}~\cite{figueiredo2002unsupervised,rosset2003boosting,tolstikhin2017adagan}, which takes the arithmetic average of meta-models' distributions, and \textit{product-of-experts}~\cite{hinton2002training,grover2018boosted}, which takes the geometric average of meta-models' distributions.

\section{Conclusion}
We propose a meta-algorithmic framework for boosting generative models by connecting meta-models in cascade. Any hidden variable meta-models can be incorporated, as long as it supports likelihood evaluation. The decomposable lower bound allows us to train these meta-models separately and greedily. Our framework can be extended to semi-supervised learning and can be integrated with multiplicative boosting. In our experiments, we first validate the effectiveness of our boosting method via density estimation and evaluating the generated samples, and then further promote the performance of some advanced models, which represent state-of-the-art methods. Finally, we compare different generative boosting methods, validating the ability of the hybrid boosting in further improving learning power of generative models.

\clearpage

\bibliographystyle{named}

\end{document}


\maketitle

\appendix
\section{Proof of the Decomposable Lower Bound}
\begin{theorem}
\label{thm:ELBO}
Let $m_i(h_{i-1}, h_i)$($1 \leq i \leq n$) be meta-models, $p_k(x, h_1, \cdots, h_k) = m_k(h_{k-1}, h_k)\prod_{i=1}^{k-1}m_i(h_{i-1}|h_i)$ be the model boosted from meta-models $m_i$($1\leq i\leq k$), $q(h_1, \cdots, h_{k-1}|x)=\prod_{i=1}^{k-1} m_i(h_i|h_{i-1})$ be the approximate posterior, then we have:
\begin{equation}
\label{eq:elbo}
\begin{aligned}
    &\mathbf{E}_D\left[\mathrm{log}p_{k}(x)\right] \geq\\  
    &\mathbf{E}_D\mathbf{E}_{q(h_1, \cdots, h_{k-1}|x)} \left[ \mathrm{log} \frac{p_{k}(x, h_1, \cdots, h_{k-1})}{q(h_1, \cdots, h_{k-1}|x)} \right]= \sum_{i=1}^{k} \mathcal{L}_k,
\end{aligned}
\end{equation}
where
\begin{equation}
\begin{aligned}
    \mathcal{L}_1 =& \mathbf{E}_D\left[\mathrm{log}m_1(x)\right]\\
    \mathcal{L}_i=&\mathbf{E}_D\mathbf{E}_{q(h_{i-1}|x)} \left[\mathrm{log}m_{i}(h_{i-1})\right]\\
    &-\mathbf{E}_D\mathbf{E}_{q(h_{i-1}|x)}\left[\mathrm{log}m_{i-1}(h_{i-1})\right] (2\leq i \leq k).
\end{aligned}
\end{equation}
\end{theorem}
\begin{proof}
Using $q(h_1, \cdots, h_{k-1}|x)$ as the approximate posterior, we have a variational lower bound for $\mathrm{log}p_{k}(x)$:
$$
\mathrm{log}p_{k}(x) \geq \mathbf{E}_{q(h_1, \cdots, h_{k-1}|x)} \left[ \mathrm{log} \frac{p_{k}(x, h_1, \cdots, h_{k-1})}{q(h_1, \cdots, h_{k-1}|x)} \right],
$$
with
$$
\begin{aligned}
\frac{p_{k}(x, h_1, \cdots, h_{k-1})}{q(h_1, \cdots, h_{k-1}|x)}=&\frac{m_{k}(h_{k-1})\prod_{i=1}^{k-1}\frac{m_i(h_{i-1}, h_i)}{m_i(h_i)}}{\prod_{i=1}^{k-1}\frac{m_i(h_{i-1}, h_i)}{m_i(h_{i-1})}}\\
=&m_1(x)\prod_{i=1}^{k-1}\frac{m_{i+1}(h_i)}{m_i(h_i)}.
\end{aligned}
$$
Thus, the lower bound is equal to:
$$
\begin{aligned}
&\mathbf{E}_{q(h_1, \cdots, h_{k-1}|x)} \left\{ \mathrm{log}\left[ m_1(x)\prod_{i=1}^{k-1}\frac{m_{i+1}(h_i)}{m_i(h_i)}\right]\right\}\\
=&\mathrm{log}m_1(x)+\sum_{i=1}^{k-1}\mathbf{E}_{q(h_i|x)}\left[\mathrm{log}\frac{m_{i+1}(h_i)}{m_i(h_i)}\right]\\
=&\mathrm{log}m_1(x)+\sum_{i=1}^{k-1} \left\{\mathbf{E}_{q(h_i|x)}\left[\mathrm{log}m_{i+1}(h_i)\right]\right.\\
&\left.-\mathbf{E}_{q(h_i|x)}\left[\mathrm{log}m_{i}(h_i)\right]\right\}.
\end{aligned}
$$
Thus,
$$
\begin{aligned}
&\mathrm{log}p_{k}(x) \geq \mathrm{log}m_1(x)\\
&+\sum_{i=1}^{k-1} \left\{\mathbf{E}_{q(h_i|x)}\left[\mathrm{log}m_{i+1}(h_i)\right]-\mathbf{E}_{q(h_i|x)}\left[\mathrm{log}m_{i}(h_i)\right]\right\}.
\end{aligned}
$$
Take the expection with respect to dataset $D$, we have
$$
\begin{aligned}
&\mathbf{E}_D\left[\mathrm{log}p_{k}(x)\right] \geq \mathbf{E}_D\left[\mathrm{log}m_1(x)\right] + \sum_{i=1}^{k-1}\\
&\left\{\mathbf{E}_D\mathbf{E}_{q(h_i|x)}\left[\mathrm{log}m_{i+1}(h_i)\right] - \mathbf{E}_D\mathbf{E}_{q(h_i|x)}\left[\mathrm{log}m_{i}(h_i)\right]\right\}\\
&=\sum_{i=1}^k \mathcal{L}_k.
\end{aligned}
$$
\end{proof}

\section{Proof of the Convergence}
\begin{theorem}
\label{thm:Convergence}
If $m_k$ reaches the best global optimum, then $\sum_{i=k+1}^j \mathcal{L}_i \leq 0$, for any $j \geq k + 1$.
\end{theorem}

\begin{theorem}
\label{thm:Convergence_necc}
If $m_k$ reaches the best global optimum, then $\mathbf{E}_D\mathbf{E}_{q(h_{k}|x)}\left[\mathrm{log}m_{k}(h_k)\right]=\mathbf{E}_{m_k(h_{k})}\left[\mathrm{log}m_{k}(h_k)\right]$.
\end{theorem}

\begin{proof}
Since $m_k$ reaches the best global optimum, $m_{k}(h_{k-1})$ is exactly the marginal distribution of $p_D(x) q(h_{k-1}|x)$ and we have
$$\mathbf{E}_D\mathbf{E}_{q(h_{k-1}|x)}\left[\mathrm{log}m_k(h_{k-1})\right]=\mathbf{E}_{m_k(h_{k-1})}\left[\mathrm{log}m_k(h_{k-1})\right].$$
For $i\geq k+1$, we have
$$
\begin{aligned}
\mathcal{L}_i=&\mathbf{E}_{m_k(h_{k-1})}\mathbf{E}_{q(h_{i-1}|h_{k-1})} \left[\mathrm{log}m_{i}(h_{i-1})\right]\\
    &-\mathbf{E}_{m_k(h_{k-1})}\mathbf{E}_{q(h_{i-1}|h_{k-1})}\left[\mathrm{log}m_{i-1}(h_{i-1})\right].
\end{aligned}
$$
Given $j\geq k+1$, let
$$p_j(h_{k-1},\cdots ,h_j)=m_j(h_{j-1}, h_j)\prod_{i=k}^{j-1}m_i(h_{i-1}|h_i).$$
and let
$$q(h_k,\cdots,h_{j-1}|h_{k-1})=\prod_{i=k}^{j-1} m_i(h_i|h_{i-1})$$
be the approximate posterior of $p_j(h_{k-1},\cdots,h_{j-1})$.

According to \autoref{thm:ELBO}, we have
$$
\begin{aligned}
&\mathbf{E}_{m_k(h_{k-1})}\left[\mathrm{log}p_j(h_{k-1})\right]\\
\geq&\mathbf{E}_{m_k(h_{k-1})}\left[\mathrm{log}m_k(h_{k-1})\right] + \sum_{i=k+1}^j \mathcal{L}_i.
\end{aligned}
$$
Since $$\mathbf{E}_{m_k(h_{k-1})}\left[\mathrm{log}p_j(h_{k-1})\right] \leq \mathbf{E}_{m_k(h_{k-1})}\left[\mathrm{log}m_k(h_{k-1})\right],$$
we have $\sum_{i=k+1}^j \mathcal{L}_i \leq 0$.
Besides, we also have 
$$
\begin{aligned}
\mathbf{E}_D\mathbf{E}_{q(h_{k}|x)}&\left[\mathrm{log}m_{k}(h_k)\right]\\
&=\mathbf{E}_{m_k(h_{k-1})}\mathbf{E}_{q(h_{k}|h_{k-1})}\left[\mathrm{log}m_{k}(h_{k})\right]\\
&=\mathbf{E}_{m_k(h_{k})}\left[\mathrm{log}m_{k}(h_k)\right].
\end{aligned}
$$
\end{proof}

\section{Semi-supervised Boosting}
\label{sec:semi}

Our meta-algorithmic framework can be extended to semi-supervised boosting. Suppose we have a large amount of labelled data and a small amount of unlabelled data, where the unlabelled dataset is denoted by $D_u$ and the labelled dataset is denoted by $D_l$. The task is to learn a conditional distribution from these data. 

To solve this task, we introduce a latent class variable $y$. The first $n-1$ meta-models doesn't have the latent class variable $y$; they are normal hidden variable models. The latent class variable $y$ only appears in the last meta-model. Thus, the joint distribution of the boosted model become
\begin{equation}
    p_n(x, y, h_1, \cdots, h_n) = m_n(h_{n-1},y, h_n)\prod_{i=1}^{n-1}m_i(h_{i-1}|h_i).
\end{equation}
Similarly, the approximation of the posterior distribution become
\begin{equation}
\label{eq:post_semi}
    q(y, h_1, \cdots, h_n|x)=m_n(y, h_n|h_{n-1})\prod_{i=1}^{n-1} m_i(h_i|h_{i-1}).
\end{equation}
For unlabelled data, we have the following decomposable lower bound,
\begin{equation}
\label{eq:elbo_unlabelled}
\begin{aligned}
    &\mathbf{E}_{D_u}\left[\mathrm{log}p_{n}(x)\right] \geq \sum_{i=1}^n \mathcal{L}_i^u, 
\end{aligned}
\end{equation}
where each decomposed term $\mathcal{L}_i^u$ is
\begin{equation}
\begin{aligned}
    \mathcal{L}_1^u=&\mathbf{E}_{D_u}\left[\mathrm{log}m_1(x)\right]\\
    \mathcal{L}_i^u=&\mathbf{E}_{D_u}\mathbf{E}_{q(h_{i-1}|x)} \left[\mathrm{log}m_{i}(h_{i-1})\right]\\
    &-\mathbf{E}_{D_u}\mathbf{E}_{q(h_{i-1}|x)}\left[\mathrm{log}m_{i-1}(h_{i-1})\right] (2\leq i \leq n).
\end{aligned}
\end{equation}

We see that these terms have no difference from the decomposed terms listed in \autoref{thm:ELBO}, except that the expectations of marginal likelihoods are taken on unlabelled dataset $D_u$. 
The proof for \autoref{eq:elbo_unlabelled} is given in C.2.

For labelled data, we have the following decomposable lower bound,
\begin{equation}
\label{eq:elbo_labelled}
\begin{aligned}
    &\mathbf{E}_{D_l}\left[\mathrm{log}p_{n}(x, y)\right] \geq \sum_{i=1}^n \mathcal{L}_i^l,
\end{aligned}
\end{equation}
where the first $n-1$ decomposed terms are
\begin{equation}
\label{eq:elbo_labelled_terms_1}
\begin{aligned}
    \mathcal{L}_1^l=&\mathbf{E}_{D_l}\left[\mathrm{log}m_1(x)\right]\\
    \mathcal{L}_i^l=&\mathbf{E}_{D_l}\mathbf{E}_{q(h_{i-1}|x)} \left[\mathrm{log}m_{i}(h_{i-1})\right]\\
    &-\mathbf{E}_{D_l}\mathbf{E}_{q(h_{i-1}|x)}\left[\mathrm{log}m_{i-1}(h_{i-1})\right] (2\leq i \leq n-1),
\end{aligned}
\end{equation}
which have no difference from the decomposed terms listed in \autoref{thm:ELBO}, except that the expectations of marginal likelihood are taken on labelled dataset $D_l$, and the last term is
\begin{equation}
\label{eq:elbo_labelled_terms_2}
\begin{aligned}
    \mathcal{L}_n^l=&\mathbf{E}_{D_l}\mathbf{E}_{q(h_{n-1}|x)} \left[\mathrm{log}m_{n}(h_{n-1},y)\right]\\
    &-\mathbf{E}_{D_l}\mathbf{E}_{q(h_{n-1}|x)}\left[\mathrm{log}m_{n-1}(h_{n-1})\right],
\end{aligned}
\end{equation}
which has a slight difference from the last term listed in \autoref{thm:ELBO}: the first part of $\mathcal{L}_n^l$ is the marginal likelihood of the ensemble of the observable variable $h_{n-1}$ and the latent class variable $y$. It is in $\mathcal{L}_n^l$ where the latent class variable $y$ is considered. The proof for \autoref{eq:elbo_labelled} is given in C.3.

We use a weighted sum of \autoref{eq:elbo_unlabelled} and \autoref{eq:elbo_labelled} to get a lower bound for the entire dataset: 
\begin{equation}
\label{eq:elbo_semi}
\begin{aligned}
    &\alpha \mathbf{E}_{D_u}\left[\mathrm{log}p_{n}(x)\right]  + \beta \mathbf{E}_{D_l}\left[\mathrm{log}p_{n}(x, y)\right] \\
    \geq  &\alpha \sum_{i=1}^n \mathcal{L}_i^u + \beta \sum_{i=1}^n\mathcal{L}_i^l\\
    = & \sum_{i=1}^n \left[\alpha \mathcal{L}_i^u + \beta \mathcal{L}_i^l\right]\\
    = & \sum_{i=1}^n \mathcal{J}_i,
\end{aligned}
\end{equation}
where $\alpha$ and $\beta$ are the weights and $\alpha + \beta =1$. The lower bound is also decomposed to $n$ terms. If we let $D$ denote the weighted combination of $D_u$ and unlabelled version of $D_l$ (i.e., the empirical distribution of $D$ is the weighted summation of $p_{D_u}$ and $p_{D_l}$: $p_D(x) = \alpha p_{D_u}(x) + \beta p_{D_l}(x)$), we can get
\begin{equation}
\begin{aligned}
    \mathcal{J}_1=&\mathbf{E}_{D}\left[\mathrm{log}m_1(x)\right]\\
    \mathcal{J}_i=&\mathbf{E}_{D}\mathbf{E}_{q(h_{i-1}|x)} \left[\mathrm{log}m_{i}(h_{i-1})\right]\\
    &-\mathbf{E}_{D}\mathbf{E}_{q(h_{i-1}|x)}\left[\mathrm{log}m_{i-1}(h_{i-1})\right] (2\leq i \leq n-1),
\end{aligned}
\end{equation}
which also have no difference from these of the lower bound in \autoref{thm:ELBO}, and
\begin{equation}
\begin{aligned}
    \mathcal{J}_n=&\alpha\mathbf{E}_{D_u}\mathbf{E}_{q(h_{n-1}|x)} \left[\mathrm{log}m_{n}(h_{n-1})\right]\\
    &+ \beta \mathbf{E}_{D_l}\mathbf{E}_{q(h_{n-1}|x)} \left[\mathrm{log}m_{n}(h_{n-1},y)\right]\\
    &-\mathbf{E}_{D}\mathbf{E}_{q(h_{n-1}|x)}\left[\mathrm{log}m_{n-1}(h_{n-1})\right].
\end{aligned}
\end{equation}
Since the first $n-1$ terms of the lower bound also have no difference from these of the lower bound in \autoref{thm:ELBO}, we can train the first $n-1$ meta-models by directly using our algorithm of boosting generative models. For the last meta-model, we train it by optimizing $\mathcal{J}_n$: only tune the parameters of $m_n$ to make $\mathcal{J}_n$ grow. The training algorithm is summarized in \autoref{alg:semitrain}.

\begin{algorithm}[h]
    \caption{Semi-Supervised Boosting}
    \label{alg:semitrain}
\begin{algorithmic}[1]
    \STATE {\bfseries Input:} unlabelled dataset $D_u$;\ \ labelled dataset $D_l$;\ \ number of meta-models $n$
    \STATE Train first $n-1$ meta-models using our algorithm of boosting generative models
    \STATE Train the last meta-model $m_n$ by optimizing $\mathcal{J}_n$ 
    \STATE Build the boosted model $p_n(x, y, h_1, \cdots, h_n) = m_n(h_{n-1},y, h_n)\prod_{i=1}^{n-1}m_i(h_{i-1}|h_i)$
    \STATE Build the approximate posterior 
    $q(y, h_1, \cdots, h_n|x)=m_n(y, h_n|h_{n-1})\prod_{i=1}^{n-1} m_i(h_i|h_{i-1})$
    \STATE Return $p_n(x, y, h_1, \cdots, h_n)$, $q(y, h_1, \cdots, h_n|x)$
\end{algorithmic}
\end{algorithm}

It is worth note that, when we restrict the number of meta-models to 2, the boosted model is exactly the stacked generative semi-supervised model~\cite{kingma2014semi}.

We don't introduce a classification loss $\mathbf{E}_{D_l}\left[-\mathbf{log}q(y|x)\right]$ ~\cite{kingma2014semi} for the boosted model. In our experiments, we will show that the performances of models with the loss term and models without the loss term are comparable on the classification task. In fact, by well balancing $\alpha$ and $\beta$, the boosted-model can have the ability for classification, which reveals that the classification task can be implemented entirely by generative methods.

\subsection{Experiments on Semi-supervised Boosting}
\label{sec:exp_semi}
This experiment is designed to show that deep generative models can perform well on the classification task without the help of any classification loss. We compare our methods (without classification loss) with the methods in Kingma \textit{et al.}~\shortcite{kingma2014semi} (with classification loss). The models covered in our experiments are listed in \autoref{tab:clssification_models}. $M_1$ and $M_2$ are models in our methods with no classification loss; $M_3$ and $M_4$ are models in the methods of Kingma \textit{et al.}~\shortcite{kingma2014semi} with classification loss.
\begin{table}[htb]
	\centering 
	\begin{tabular}{|c|c|c|}  
		\hline
		  &has classification loss? & \#meta-models\\
		\hline
        $M_1$ & Y & 1 \\
        \hline
        $M_2$ & Y & 2\\
        \hline
        $M_3$ & N & 1\\
        \hline
        $M_4$ & N & 2\\
        \hline
	\end{tabular}
	\caption{The boosted models covered in our experiments of semi-supervised learning.}
	\label{tab:clssification_models}
\end{table}

We do experiments under different amount of labeled data: we randomly pick 100, 600, and 1000 and 3000 labelled data from mnist training data, where each class has the same number of labelled data. The results are shown in \autoref{tab:clssification}. The classification accuracy for models with and without the classification loss are comparable. Under some cases, the models without loss terms can outperform models with loss terms ($M_1$ outperforms $M_3$ on any number of labelled data, while they have the same number of meta-models). This result suggests us that the generative model can learn how to classify without telling it how to classify.

\begin{table}[htb]
	\centering 
	\begin{tabular}{|c|c|c|c|c|}  
		\hline
		$|D_l|$&$M_1$&$M_2$&$M_3$&$M_4$\\
		\hline
        100 & 94.32 & 95.20 & 88.03 & 96.67 \\
        \hline
        600 & 96.04 & 95.74 & 95.06 & 97.41 \\
        \hline
        1000 & 97.38 & 96.10 & 96.40 & 97.60 \\
        \hline
        3000 & 97.52 & 96.57 & 96.08 & 97.82 \\
        \hline
	\end{tabular}
	\caption{The classification accuracy on mnist test dataset for models with and without the classification loss. $|D_l|$ is the number of labelled data.}
	\label{tab:clssification}
\end{table}

\subsection{Proof of the Decomposable Lower Bound for Unlabelled Data}
\begin{proof}
If we view $h_n$ and $y$ as a whole $H_n$, the joint distribution of the boosted model can be written as
$$
    p_n(x, h_1, \cdots, H_n) = m_n(h_{n-1},H_n)\prod_{i=1}^{n-1}m_i(h_{i-1}|h_i),
$$
and the approximation of the posterior distribution can be written as 
$$
     q(h_1, \cdots, H_n|x)=m_n(H_n|h_{n-1})\prod_{i=1}^{n-1} m_i(h_i|h_{i-1}).
$$
Now, we can directly leverage \autoref{thm:ELBO} to get the decomposable lower bound:
\begin{equation}
\begin{aligned}
    &\mathbf{E}_{D_u}\left[\mathrm{log}p_{n}(x)\right] \geq \sum_{i=1}^n \mathcal{L}_i^u, 
\end{aligned}
\end{equation}
where each decomposed term $\mathcal{L}_i^u$ is
\begin{equation}
\begin{aligned}
    \mathcal{L}_1^u=&\mathbf{E}_{D_u}\left[\mathrm{log}m_1(x)\right]\\
    \mathcal{L}_i^u=&\mathbf{E}_{D_u}\mathbf{E}_{q(h_{i-1}|x)} \left[\mathrm{log}m_{i}(h_{i-1})\right]\\
    &-\mathbf{E}_{D_u}\mathbf{E}_{q(h_{i-1}|x)}\left[\mathrm{log}m_{i-1}(h_{i-1})\right] (2\leq i \leq n).
\end{aligned}
\end{equation}
\end{proof}

\subsection{Proof of the Decomposable Lower Bound for Labelled Data}
\begin{proof}
The marginal likelihood of labelled data can be broken down into two parts:
$$
\mathbf{E}_{D_l}\left[\mathrm{log}p_{n}(x, y)\right] =\mathbf{E}_{D_l}\left[\mathrm{log}p_{n}(x| y)\right]+\mathbf{E}_{D_l}\left[\mathrm{log}m_{n}( y)\right].
$$
We derive a lower bound for the first part $\mathbf{E}_{D_l}\left[\mathrm{log}p_{n}(x| y)\right]$. Since the boosted model conditioned on $y$ is
$$
p_n(x, h_1, \cdots, h_n|y) = m_n(h_{n-1}, h_n|y)\prod_{i=1}^{n-1}m_i(h_{i-1}|h_i),
$$
and the approximation of the posterior conditioned on $y$ is 
$$
q(h_1, \cdots, h_n|x, y)=m_n(h_n|h_{n-1}, y)\prod_{i=1}^{n-1} m_i(h_i|h_{i-1}).
$$
By leveraging \autoref{thm:ELBO}, we can derive a lower bound for $\mathbf{E}_{D_l}\left[\mathrm{log}p_{n}(x| y)\right]$:
$$
\begin{aligned}
\mathbf{E}_{D_l}\left[\mathrm{log}p_{n}(x| y)\right] \geq  &\sum_{i=1}^{n-1}\mathcal{L}_i^l+\hat{\mathcal{L}_n^l},
\end{aligned}
$$
where 
$$
\begin{aligned}
    \mathcal{L}_1^l=&\mathbf{E}_{D_l}\left[\mathrm{log}m_1(x)\right]\\
    \mathcal{L}_i^l=&\mathbf{E}_{D_l}\mathbf{E}_{q(h_{i-1}|x)} \left[\mathrm{log}m_{i}(h_{i-1})\right]\\
    &-\mathbf{E}_{D_l}\mathbf{E}_{q(h_{i-1}|x)}\left[\mathrm{log}m_{i-1}(h_{i-1})\right] (2\leq i \leq n-1),
\end{aligned}
$$
and  
$$
\begin{aligned}
\hat{\mathcal{L}_n^l}=&\mathbf{E}_{D_l}\mathbf{E}_{q(h_{n-1}|x)} \left[\mathrm{log}m_{n}(h_{n-1}|y)\right]\\&-\mathbf{E}_{D_l}\mathbf{E}_{q(h_{n-1}|x)}\left[\mathrm{log}m_{n-1}(h_{n-1})\right].
\end{aligned}
$$
Then we have
$$
\begin{aligned}
    \mathbf{E}_{D_l}\left[\mathrm{log}p_{n}(x, y)\right] =&
    \mathbf{E}_{D_l}\left[\mathrm{log}p_{n}(x| y)\right]+
    \mathbf{E}_{D_l}\left[\mathrm{log}m_{n}( y)\right]\\
   \geq  &\sum_{i=1}^{n-1}\mathcal{L}_i^l + \hat{\mathcal{L}_n^l}+\mathbf{E}_{D_l}\left[\mathrm{log}m_{n}( y)\right],
\end{aligned}
$$
and
$$
\begin{aligned}
    &\hat{\mathcal{L}_n^l}+\mathbf{E}_{D_l}\left[\mathrm{log}m_{n}( y)\right]\\
    =&\mathbf{E}_{D_l}\mathbf{E}_{q(h_{n-1}|x)} \left[\mathrm{log}m_{n}(h_{n-1}|y)\right]+\mathbf{E}_{D_l}\left[\mathrm{log}m_{n}( y)\right] \\
    &-\mathbf{E}_{D_l}\mathbf{E}_{q(h_{n-1}|x)}\left[\mathrm{log}m_{n-1}(h_{n-1})\right]\\
    =&\mathbf{E}_{D_l}\mathbf{E}_{q(h_{n-1}|x)} \left[\mathrm{log}m_{n}(h_{n-1},y)\right]\\
    &-\mathbf{E}_{D_l}\mathbf{E}_{q(h_{n-1}|x)}\left[\mathrm{log}m_{n-1}(h_{n-1})\right]\\
    =&\mathcal{L}_n^l.
\end{aligned}
$$
Thus, we have $\mathbf{E}_{D_l}\left[\mathrm{log}p_{n}(x, y)\right] \geq \sum_{i=1}^{n}\mathcal{L}_i^l$.

\end{proof}

\section{Architectures of Meta-Models}
The architectures of VAEs, ConvVAEs, IWAEs and LVAEs are given in this part.
\subsection{Architectures of VAEs}
All VAEs have two deterministic hidden layers for both generation, and inference and we add batch normalization layers \cite{ioffe2015batch,sonderby2016ladder} after deterministic hidden layers. The dimension of deterministic hidden layers is set to 500 and 2500, and the dimension of stochastic hidden variables is set to 20 and 100, for experiments on mnist and celebA respectively. 

\subsection{Architectures of ConvVAEs}
The ConvVAE has one 500-dimensional deterministic hidden layer and one 50-dimensional stochastic hidden variable, with four additional convolutional layers \cite{lecun1998gradient}. All convolutional layers have a kernel size of $4 \times 4$ and a stride of 2. Their channels are 32, 64, 128 and 256 respectively. We add batch normalization layers after deterministic hidden layers.

\subsection{Architectures of IWAEs}
The IWAE has two 500-dimensional deterministic hidden layers and one 50-dimensional stochastic hidden variable. The number of importance sampling is set to 5 and 10.

\subsection{Architectures of LVAEs}
The LVAE has four 1000-dimensional deterministic hidden layers and two 30-dimensional stochastic hidden variables. We add batch normalization layers after deterministic hidden layers.

\bibliographystyle{named}
\bibliography{ijcai19}